\def\year{2022}\relax
\documentclass[letterpaper]{article} 
\usepackage{aaai22}  
\usepackage{times}  
\usepackage{helvet}  
\usepackage{courier}  
\usepackage[hyphens]{url}  
\usepackage{graphicx} 
\usepackage{natbib}  
\usepackage{caption} 
\DeclareCaptionStyle{ruled}{labelfont=normalfont,labelsep=colon,strut=off} 
\usepackage{algorithm}
\usepackage{amsmath}
\usepackage{multirow}
\usepackage{amsthm}
\usepackage{makecell}
\DeclareMathOperator*{\argmin}{arg\,min}
\usepackage[flushleft]{threeparttable}
\usepackage{algorithm,algcompatible}%

\newtheorem{theorem}{\textbf{Theorem}}

\DeclareMathOperator*{\argmax}{arg\,max}
\newtheorem{definition}{\textbf{Definition}}
\newtheorem{Property}{\textbf{Property}}

\algnewcommand\algorithmicreturn{\textbf{return}}
\algnewcommand\RETURN{\algorithmicreturn}
\algnewcommand\algorithmicprocedure{\textbf{procedure}}
\algnewcommand\PROCEDURE{\item[\algorithmicprocedure]}%
\algnewcommand\algorithmicendprocedure{\textbf{end procedure}}
\algnewcommand\ENDPROCEDURE{\item[\algorithmicendprocedure]}%
\algnewcommand{\algvar}[1]{{\text{\ttfamily\detokenize{#1}}}}
\algnewcommand{\algarg}[1]{{\text{\ttfamily\itshape\detokenize{#1}}}}
\algnewcommand{\algproc}[1]{{\text{\ttfamily\detokenize{#1}}}}
\algnewcommand{\algassign}{\leftarrow}

\usepackage{newfloat}
\usepackage{listings}
\floatstyle{ruled}
\newfloat{listing}{tb}{lst}{}
\floatname{listing}{Listing}
\usepackage{color}
\nocopyright
\title{LaPLACE: Probabilistic Local Model-Agnostic Causal Explanations}

\title{LaPLACE: Probabilistic Local Model-Agnostic Causal Explanations}
\author {
    Sein Minn    
}
\affiliations {
    \{sein.minn.cs\}@gmail.com
}

\usepackage{bibentry}

\begin{document}

\maketitle

\begin{abstract}
Machine learning models have undeniably achieved impressive performance across a range of applications. However, their often perceived "black box" nature, and lacking transparency in decision-making, have raised concerns about understanding their predictions. To tackle this challenge, researchers have developed methods to provide explanations for machine learning models. One attention-gaining approach involves causal graphs, which offer a coherent and interpretable representation of explanations. By tapping into causal relationships, these explanations empower humans to comprehend the factors steering the model's predictions, thereby fostering trust in the decision-making process.

In this paper, we introduce LaPLACE-explainer, designed to provide probabilistic cause-and-effect explanations for \textit{any} classifier operating on tabular data, in a human-understandable manner. The LaPLACE-Explainer component leverages the concept of a Markov blanket to establish statistical boundaries between relevant and non-relevant features automatically. This approach results in the automatic generation of optimal feature subsets, serving as explanations for predictions. Importantly, this eliminates the need to predetermine a fixed number ($N$) of top features as explanations, enhancing the flexibility and adaptability of our methodology. Through the incorporation of conditional probabilities, our approach offers probabilistic causal explanations and outperforms \textbf{LIME} and \textbf{SHAP} (well-known model-agnostic explainers) in terms of local accuracy and consistency of explained features. \textbf{LaPLACE}'s soundness, consistency, local accuracy, and adaptability are rigorously validated across various classification models. Furthermore, we demonstrate the practical utility of these explanations via experiments with both simulated and real-world datasets. This encompasses addressing trust-related issues, such as evaluating prediction reliability, facilitating model selection, enhancing trustworthiness, and identifying fairness-related concerns within classifiers. Our source code and data for replicating our experiments are available at: \textbf{LaPLACE}: https://github.com/Simon-tan/LaPLACE.git.
\end{abstract}

\section{INTRODUCTION}

Machine Learning (ML) models have become indispensable across various domains due to their remarkable performance. Whether humans are direct users of machine learning classifiers or deploy models within other applications, a crucial concern arises: comprehending the models' behavior and establishing trust in their predictions. If users lack an understanding of a model's predictions, their acceptance and utilization are hindered. Hence, prioritizing trust-building between humans and machine learning systems is crucial. Deploying machine learning models without transparency regarding their decision-making factors and underlying characteristics can lead to significant challenges in decision-making systems today.

While simple models are inherently interpretable, making their reasoning processes transparent, this becomes more challenging with complex models like ensemble methods or deep networks. Directly interpreting these intricate models may not provide a clear understanding of their decision-making. To address this, we propose LaPLACE-Explainer, a simple probabilistic graphical model-based interpretation grounded in probability theory. This approach ensures transparent reasoning, as the model's predictions are rooted in well-defined probabilistic principles, explicitly accounting for prediction uncertainty – particularly crucial in risk-sensitive domains. LaPLACE-Explainer utilizes a Bayesian network framework to provide explanations for classifier predictions, considering dependencies among explained features and offering causal explanations that go beyond additive feature attribution methods. By leveraging probabilistic graphical models, our approach bridges the gap between complex model behavior and human understanding, enhancing trust and interpretability with cause-and-effect explanations in machine learning systems.
\begin{figure*}[h]
\centering
  \includegraphics[width=10cm,height=4cm]{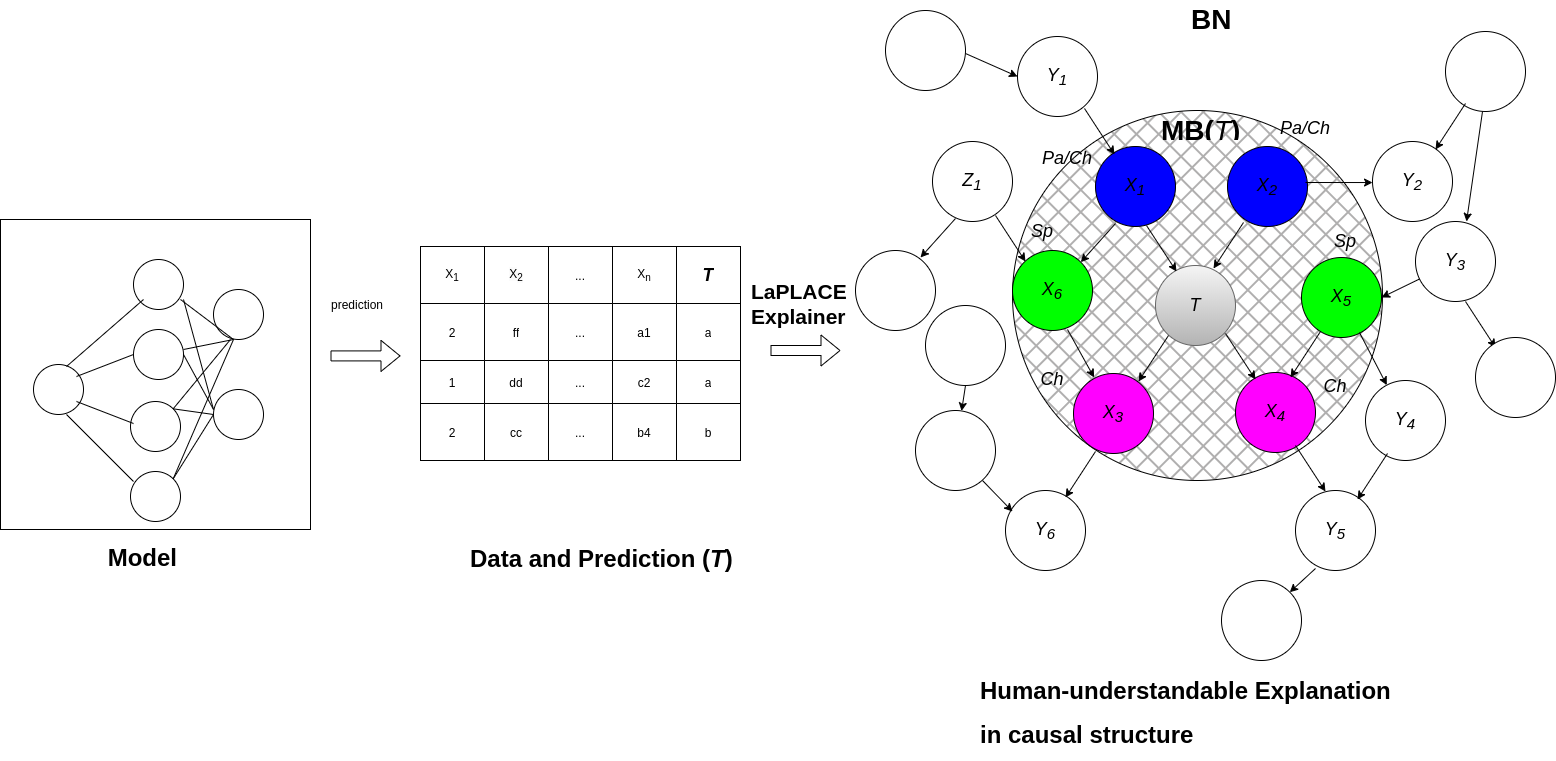}
  \caption{An illustration of an explanation within a Markov blanket $\textbf{MB}(T)$. In this figure, ${X_1 , X_2}$ are indicated as $T$'s parents $\textbf{Pa}(T)$ in blue. Similarly, ${X_3 , X_4}$ represent $T$'s children, $\textbf{Ch}(T)$ are shown in purple. Additionally, ${X_5 , X_6}$ signify $T$'s spouses $\textbf{Sp}(T)$, and are depicted in green. It provides a clear overview of the optimal feature subset within $\textbf{MB}(T)$ are associated with the target $T$, showing their causal relationships and the reset features outside of $\textbf{MB}(T)$ are non-relevant ones.}
\end{figure*}

\section{CAUSAL EXPLANATIONS}

 Interpretability helps users, stakeholders, and even regulators to understand the decision-making process of the model, build trust, and gain insights into the underlying mechanisms driving predictions. Probabilistic Graphical Models (PGMs) are built on probability theory, making their reasoning process transparent and the uncertainty associated with each prediction is explicitly accounted for, which is crucial in risk-sensitive domains. The model parameters in PGMs often have a direct probabilistic interpretation, making them more human-readable than some other explanation models. However, the level of interpretability in PGMs may vary depending on the complexity of the model and the size of the graph. In some cases, the graphical representation can become complex and challenging to interpret, especially in domains with a large number of variables and dependencies.

\subsection{Causal Model as Explainer}
We introduce a novel approach called Probabilistic Local model-Agnostic Causal Explanation (LaPLACE), utilizing a Bayesian network (BN) as a probabilistic graphical model. LaPLACE aims to establish an interpretable model within an interpretable representation that maintains local faithfulness to the classifier. Bayesian networks, as effective probabilistic graphical models, excel in modeling causality through joint probability distributions (JPDs) and supporting causal inference. By utilizing directed edges to signify influence direction, BNs adeptly capture cause-and-effect relationships, enhancing interpretability, and systematically representing variable interdependencies.

The computational complexity associated with structure learning in Bayesian networks (BNs) poses a significant challenge. The problem of identifying the optimal network structure has been recognized as NP-hard, rendering exact solutions increase the complexity with a higher number of variables involved~\cite{chickering1994learning,fu2014towards,2014Structure,2015Accelerating}. Hence, our study centers on initiating the Markov blanket of the target variable as the foundation of our explanation model. This involves treating the target node (prediction) and attribute nodes (features) equitably, followed by the utilization of a comprehensive Bayesian network to offer probabilistic explanations. This novel approach extends the capabilities of Bayesian network classifiers, enabling comprehensive explanations for diverse classifiers by uncovering underlying causal structures and incorporating causal reasoning, our proposed method enhances the interpretability of classifier predictions.

\subsubsection{Causal Reasoning based Explanation}

Bayesian networks provide a robust framework to model relationships between evidence and target variables, effectively handling uncertainty in knowledge representation. By considering direct causal influences from parent nodes, BNs enable comprehensive analysis of dependencies and influences inherent in the network. Effect variables' occurrence is influenced by direct causes represented by parent variables. Leveraging cause variable information, predictions and inferences about effect variables can be estimated. Conditional probability tables associated with target variables in BNs facilitate this prediction process. These tables capture conditional probabilities of target variables given their parent variables, enabling probability estimation and predictions based on available cause variable data.

In the context of explanation discovery in Bayesian networks, the aim is to infer probable causes contributing to an observed effect variable (predictions). Utilizing causal relationships encoded in the network, backward reasoning is employed to identify potential causes consistent with observed effects. This sheds light on underlying factors influencing outcomes, offering explanatory insights. Conditional probability is a cornerstone of Bayesian networks, facilitating uncertainty reasoning and causal understanding between variables~\cite{taroni2004general,minn2014efficient,2015Accelerated}. For a specific variable, its direct causal influences stem from its parent nodes. Bayesian networks excel at identifying crucial influential factors, thereby enhancing our comprehension of complex systems. This ability to model and reason about uncertainty and relationships between variables contributes to informed decision-making across various domains, thereby offering a systematic framework for explanations.

\subsection{Contribution}
Our specific contribution focuses on addressing the following critical questions in the context of explanation discovery:
\begin{enumerate}
    \item Faithfulness: We explore whether the explanations accurately represent how the model makes its predictions and captures its decision-making process.
    \item Trustworthiness: We seek to bridge the gap between complex ML models and users' comprehension by providing interpretable and meaningful cause-and-effect explanations.
    \item Fairness: By analyzing their causal explanations, we seek insights into the model's potential biases and facilitate a comprehensive evaluation of its performance.
\end{enumerate}

In addressing these challenges, our work contributes to the advancement of explainable artificial intelligence (XAI), striving to make machine learning models more interpretable, trustworthy, and fair. By proposing potential solutions, we aim to empower users with the tools to comprehend, trust, and assess machine learning models effectively.

\section{LaPLACE Explainer}\label{sec:laplace}
\subsection{Theoretical Basis}
A Bayesian network (BN) consists of a directed acyclic graph (DAG) $G$ with nodes representing random variables $\mathbf{U} = {(X_1, \ldots, X_n)}$. Each node $X_i$ in the graph is associated with conditional probability parameters $\theta$, denoting $P(X_i | \textbf{Pa}(X_i))$, which signifies the probability of $X_i$ given its parent nodes $\textbf{Pa}(X_i)$. The joint probability distribution for all variables in the network can be factorized as 
\begin{equation}\label{equ:cpd}
P(X_1,X_2,..,X_n) = \Pi_{i=1}^n P(X_i|\textbf{Pa}(X_i))
\end{equation}

\begin{definition}\label{df:FC}
\textbf{( Faithfulness Assumption )} A Bayesian Network $G$ and a joint distribution $P$ are faithful to one another iff. every (conditional) independence relation entailed by $P$ is also present in $G$.
\end{definition}

In a Bayesian network structure (DAG $G$), we define paths between nodes $X$ and $Y$ as sequences of nodes connected by directed arcs, ensuring no node repeats within the sequence. If there exists a directed path from node $X$ to node $Y$, we label $X$ as an ancestor of $Y$, and $Y$ as a descendant of $X$. Nodes $X$, $Y$, and $Z$ form a v-structure if $Z$ has two incoming arcs from $X$ and $Y$, but $X$ and $Y$ are not adjacent. This v-structure is represented as $X \rightarrow Z \leftarrow Y$, where $Z$ is a collider. Throughout the paper, a variable set by bold uppercase (e.g. $\textbf{X}$ ), a single variable by non-bold upper character (e.g. $X$), and their assignments by lowercase (e.g. $\textbf{X}=\textbf{x}, X=x$ ). The notation $X \perp Y \mid \mathbf{Z}$ denotes conditional independence between $X$ and $Y$ given $\mathbf{Z}$, while $X \not\perp Y \mid \mathbf{Z}$ indicates their conditional dependency.

\begin{definition}\label{df:CC}
\textbf{(Conditional independence)} Variables X and Y are conditionally independent given the set of variables
\textbf{Z} iff. $P (X \mid Y, \textbf{Z}) = P (X \mid \textbf{Z})$, denoted as $X \perp Y \mid \textbf{Z}$.
\end{definition}

\subsection{Human-understandable Causal Explanation}

In the context of LaPLACE explanations, the process of identifying the Markov blanket ($\textbf{MB}(T)$) for the target variable holds significant importance. While the determination of $\textbf{MB}(T)$ is straightforward when the Bayesian network (BN) structure is known, challenges arise when learning the BN's structure from data, a task known to be computationally expensive and NP-hard~\cite{chickering1994learning}, especially in scenarios involving high-dimensional feature spaces~\cite{de2011efficient}. This complexity underscores the difficulty of accurately establishing $\textbf{MB}(T)$ when working with real-world datasets, emphasizing the need for efficient structure learning algorithms. Koller and Sahami (K\&S) introduced an approximate and extensible algorithm for MB induction in 1996, mitigating this challenge and sparking subsequent research advancements in the domain~\cite{koller1996toward,2016Accelerating,2016Algorithm}. This has led to the recognition of the Markov blanket's effectiveness across diverse domains.

In this section, our exploration delves into the concept of the Markov blanket within the specific context of feature subset selection and explanation discovery. We aim to shed light on the pivotal role that the Markov blanket plays in facilitating human-understandable causal interpretations of the selected features.

\begin{definition}\label{df:mc}
\textbf{( Markov Assumption )} A node in a BN is independent of its non-descendant nodes, given its parents.
\end{definition}

\begin{equation}    
\forall X \in \mathbf{U} \setminus \mathbf{MB}(T) \setminus \{T\}, X \perp T|\mathbf{MB}(T)
\end{equation}
The Markov assumption, as defined in Definition \ref{df:mc}, underpins the concept of the Markov blanket. The minimum Markov blanket, known as the Markov boundary, embodies the set of nodes that are directly connected to the target variable $T$.

K\&S assert that obtaining all necessary features for a complete determination of the target variable is impractical, given the uncertainty of feature relevance and potential redundancy. Consequently, prediction fundamentally involves determining the posterior probability, $P(T | \mathbf{U})$. To select a subset $\mathbf{S} \subseteq \mathbf{U}$ capable of representing $\mathbf{U}$ fully or substantially, an optimal solution would minimize the disparity between the distributions $P(T | \mathbf{S})$ and $P(T | \mathbf{U})$, expressed as $\argmin_{\mathbf{S} \subseteq \mathbf{U}} \mathbf{D}(P(T|\mathbf{U}), P(T|\mathbf{S}))$.

\begin{theorem} \label{the:1}
Given the faithfulness assumption, 
\begin{enumerate}
    \item $\mathbf{MB}(T)$ is unique and only contains $\mathbf{Pa}(T)$,$\mathbf{Ch}(T)$ and $\mathbf{Sp}(T)$; 
    \item $ \forall X \in \mathbf{V} \backslash \mathbf{MB}(T) \backslash \{T\}, I_{\mathbf{G}} (X,T|\mathbf{MB}(T))$
    
\end{enumerate}
\end{theorem}
Under the causal Markov assumption, each variable within a BN exhibits independence from its ancestors, given the values of its parent nodes~\cite{pearl1988probabilistic}. K\&S further establishes that the Markov blanket (MB) represents the optimal feature subset for prediction, leveraging information entropy and cross-entropy. Consequently, $P(T | \mathbf{MB}(T))$ is closer to $P(T | \mathbf{U})$ than any other subset $P(T|\mathbf{S})$. Similarly, Singh and Provan's work in the same year~\cite{singh1996efficient} follows a comparable approach, with algorithms in this category founded on MB's definition and Theorem \ref{the:1}.

The $\mathbf{MB}(T)$ can be determined through various MB induction algorithms, including K\&S~\cite{koller1996toward}, GSMB~\cite{margaritis1999bayesian}, IAMB and its variants~\cite{tsamardinos2003time,tsamardinos2003algorithms,yaramakala2005speculative,borboudakis2019forward}, PC-MB~\cite{pena2007towards}, IPC-MB~\cite{fu2008fast}, among others. However, we opt for the more time and data-efficient IPC-MB algorithm~\cite{fu2008fast,minn2014efficient,yu2020causality} in the LaPLACE explainer.

IPC-MB relies on RecognizePC to ascertain parent-child relationships. Notably, IPC-MB's RecognizePC employs a backward selection approach, in contrast to the forward selection method used in PC-MB. Initially, all nodes are assumed to be adjacent to node $T$, progressively eliminating false positive connections by performing conditional independence (CI) tests (e.g. $ {\chi}^2 $ test with $\alpha=0.001$)  among nodes (variables) through systematic iterations. It employs a local criterion to identify conditional independence relationships by examining smaller subsets of variables at a time, making them more data-efficient. A node is flagged as a false positive if a cut set is identified. The cut set begins as an empty set and expands iteratively, allowing the removal of false positives using minimal cut set expansion. This iterative process efficiently identifies and rectifies inaccurate parent-child relationships in the BN structure. IPC-MB can distinguish parents and children from spouses within $\mathbf{MB}(T)$, identify some children from parents/children, and partial $v$-structures. Nonetheless, we proceeded to reapply a structure learning algorithm to the chosen features within $\mathbf{MB}(T)$, adhering to the guidelines established in the BN package of Weka. As a result, we produce the explanation as a causal graph for the given data.

\subsection{Probabilistic Causal Explanation}

Completing the qualitative structure of a $\mathbf{MB}(T)$ involves the intricate process of identifying relevant variables and establishing directed edges between them, ultimately creating a DAG. This comprehensive approach not only enhances the comprehension of the underlying relationships but also contributes to the model's predictive capabilities. While our primary emphasis is on furnishing causal explanations, it is paramount to evaluate and understand the connections within the generated graph.

Transitioning to the quantitative dimension of the network, we briefly delve into the fundamental principles of the Bayes theorem. According to this theorem, a variable, given knowledge of its parent nodes, retains independence from all other variables that do not belong to its descendant nodes within the Bayesian Network (BN). When these conditional relationships, as indicated by the structure of the BN, hold true for a set of variables, the computation of the joint probability distribution can be achieved by multiplying the specific conditional probabilities, as exemplified in Equation~\ref{equ:cpd}. Notably, this mathematical framework adheres to the chain rule, which is applicable to three fundamental types of connections within the network: serial (1), diverging (2), and converging (3).

\begin{enumerate}
    \item  X $\rightarrow$ Z $\rightarrow$ Y : $P (X, Z, Y) =  P (X) · P (Z | X) · P (Y | Z)$
    \item  X $\leftarrow$ Z $\rightarrow$ Y :  $P (Z, X, Y) = P (Z) · P (X | Z) · P (Y | Z)$
    \item  X $\rightarrow$ Z $\leftarrow$ Y : $P (X, Y, Z) = P (X) · P (Y) · P (Z | X,Y)$
\end{enumerate}

\begin{figure}[h]
\centering
  \includegraphics[width=6cm,height=3.5cm]{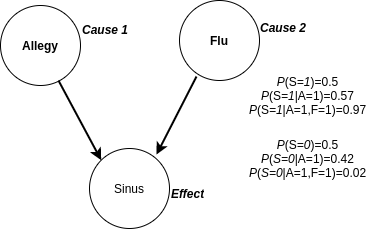}
  \caption{ An example of BN implying three variables $(A, F, S)$ which contains one child node $S$ and two parent nodes $A,F$ in a converging connection, probabilistic reasoning enables accounting for direct causal influences originating from parent nodes, impacting a specific variable $S$. The network's graphical representation and conditional probabilities facilitate understanding the probabilistic relationships and dependencies among variables.}\label{fig_mle}
\end{figure}

Quantitatively specifying a BN involves assigning node probabilities to each variable. While manually tuning parameters can be intricate, learning from data often yields enhanced performance. This task is proven to be NP-hard~\cite{barahona1982computational}. A prevalent approach for parameter estimation is maximum-likelihood estimation, entailing the maximization of the likelihood of observed data based on the given parameters~\cite{spirtes2000causation}. To illustrate, consider the example depicted in Figure~\ref{fig_mle}, encompassing three binary random variables: $A, F,$ and $S$, where $S$ is influenced by two parent variables, $A$ and $F$. To estimate the probability $P ( A, F, S) $. The maximum-likelihood estimation that maximizes the likelihood of observing the data given these parent variables is

\begin{align}
    \theta_{s| i,j} & = P (S=1|A=i,F=j) \\
                    & = \frac{ \Sigma_{k=1}^K \delta (a_k=i, f_k=j, s_k=1)}{\Sigma_{k=1}^K \delta(a_k=i,f_k=j)}
\end{align}

where $K$ represents the count of fully observed training samples and $\delta(x)$ equals 1 if $x$ is true and 0 otherwise in the case of binary class problem, the framework yields local maximum likelihood estimates. Even when the target variable $T$ remains unobserved, it accommodates scenarios where other nodes within the $\mathbf{MB}(T)$ possess observed values. By employing maximum likelihood estimation (MLE), the overall global likelihood can be disintegrated into distinct terms, with each term corresponding to a variable within the network. The inference task transforms into an MPE (Most Probable Explanation) task, involving the determination of the marginal probability $P(T = t | \mathbf{MB}(T) \subseteq U)$. MPE is employed to compute the most probable state of a subset of variables within the network, given observed values of other variables during the explanation of $\mathbf{MB}(T)$. These individual terms encapsulate local likelihood functions, quantifying the predictive fidelity of each variable in light of its parent variables. The inference task transforms into an MPE (Most Probable Explanation) task, involving the determination of the marginal probability $P(T = t | \mathbf{MB}(T) \subseteq U)$. MPE is employed to compute the most probable state of a subset of variables within the network, given observed values of other variables during the explanation of $\mathbf{MB}(T)$. This comprehensive understanding of connections and their probabilistic implications contributes to a holistic interpretation of probabilistic causal explanation by employing BN' parameters on $\mathbf{MB}(T)$. While our focus resides in specific facets of causal explanations within this study, those intrigued can delve into comprehensive resources like ~\cite{pearl1988probabilistic, borgelt2002graphical, koller2009probabilistic} for a more profound comprehension of Bayesian network learning and its associated dimensions.

\begin{figure*}[h]
\centering
  \includegraphics[width=18cm,height=6cm]{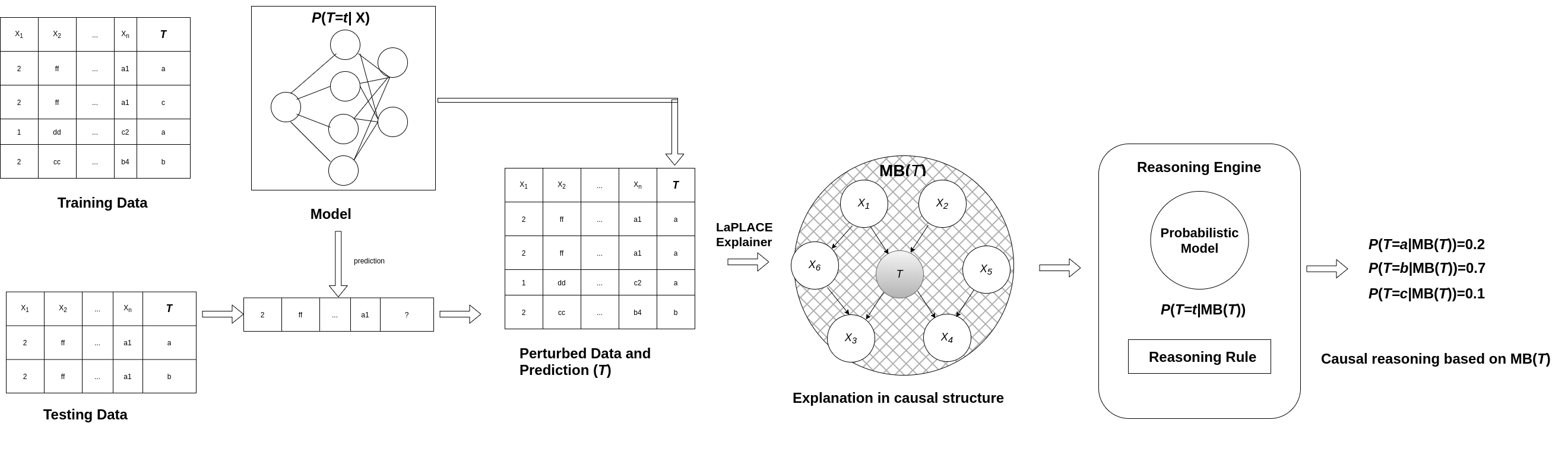}
  \caption{LaPLACE-Explainer's architecture encompasses generating perturbed data from the training distribution and utilizing $\mathbf{MB}(T)$ for significant feature extraction and causal structure identification. It then learns causal parameters, enabling comprehensive probabilistic causal explanations for predictions.}
\end{figure*}

\begin{algorithm}
\caption{Generating LaPLACE explanation}
\label{EXstream2}
\begin{algorithmic}
\raggedright

  \REQUIRE \textrm{ Classifier$f$, Instance $x$}
  \REQUIRE \textrm{ Training data $TD$, Sampling size $N$ }
  \ENSURE $\mathbf{MB}(T)$, $\mathbf{BN}$\\
   \hspace{0.9cm}$TS \gets $\textrm{ Compute frequency distribution ($TD$)}\\
   \hspace{0.9cm}  \textrm{Perturbed data }$ \gets $ \textrm{Random Perturb ($x,TS,N$)}\\
   \hspace{0.9cm}  \textrm{Perturbed data }$ \gets $ \textrm{Predict $T$ ($f,$ Perturbed data)}\\
   \hspace{0.9cm}  $ \mathbf{MB}(T)\gets$ \textrm{IPC-MB (Perturbed data)}\\
   \hspace{0.9cm}  $ \mathbf{BN}\gets$ \textrm{Structure Learning ($\mathbf{MB}(T)$)}\\
  \hspace{0.9cm}  $ \mathbf{BN}\gets$ \textrm{Parameter Learning($\mathbf{BN}$)}\\
  
\RETURN{}   $ \mathbf{MB}(T)$, $\mathbf{BN}$
\end{algorithmic}
\end{algorithm}

\subsection{Properties of Probabilistic Causal Explanation}

\begin{Property}\label{df:LAP}
\textbf{(Local accuracy)} When approximating the original model $f$ for a specific input $x$, achieving local accuracy in the explanation model necessitates that the explanation model must, at a minimum, match the output of $f$ for the perturbed input $x'$. 
\end{Property}
 . 
\begin{align}
    f(x) & = g(x')\\
         & = \argmax_t P(T|X=x') \\
         & = \argmax_t P(T|\mathbf{Pa}(T)) \Pi_{i=1}^{|U|} P(X_i|\mathbf{Pa}(X_i)=x')
\end{align}

The explanation model $g(x')$ is approximately equivalent to the original model $f(x)$ when the input $x$ is substituted with the perturbed version $x'$. The value $t$ that maximizes the marginal probability is the prediction of $g(x)$, which matches the prediction of $f(x)$. This ensures that the explanation model captures the essential behavior of the original model around the specific input instance $x$ and faithfully represents its prediction process.

\begin{Property}\label{df:ConsistencyP}

\textbf{(Consistency)}  When we modify the inputs, the impact of a particular feature on the model's output should remain consistent, and any increase in its influence should be reflected in the explanation model's attributions. 

\end{Property}

\begin{equation}\label{equ:consistency}
\argmax_t P(T = t|X=x)=\argmax_t P(T = t|\mathbf{MB}(T)=z')
\end{equation}

Where $\mathbf{MB}(T) \subseteq U$ signifies the optimal feature subset for predicting variable $T$, and $z'$ vectors denote the values corresponding to features within a subset of the optimal feature set within vector $x$. 

This property ensures that the explanation model has consistent and dependable interpretations, even when the original model undergoes modifications. This stability and reliability in interpretation contribute to the robustness and enduring utility of the explanation framework, reinforcing its practical value in dynamic and evolving scenarios.\footnote{Our source code and data for replicating our experiments are available at: \textbf{LaPLACE}: https://github.com/Simon-tan/LaPLACE.git }

\section{EXPERIMENTS}
In this section, we present our experiments to evaluate the utility of explanations in trust-related tasks.

\subsection{Experiment Setup}

We assessed the effectiveness of the LaPLACE explainer by comparing it with the Linear LIME~\cite{ribeiro2016should}\footnote{\textbf{LIME}:https://github.com/marcotcr/lime} and Kernel SHAP ~\cite{lundberg2017unified}\footnote{\textbf{SHAP}:https://github.com/shap/shap} approximation techniques. Our evaluation included accuracy comparisons with SHAP and LIME. Additionally, we conducted a consistency study to compare the $\mathbf{MB}(T)$ explanations provided by LaPLACE with alternative feature importance allocations from LIME and SHAP.

\subsection{Datasets}
\paragraph{Real world network}
We generated a dataset comprising 10,000 samples from the real-world Bayesian network,  A Logical Alarm Reduction Mechanism (ALARM) by employing the data generation technique available in the Weka. ALARM serves as a network designed to monitor patients in intensive care settings. It was initially introduced by ~\cite{beinlich1989alarm} and comprises 37 nodes, each with two to four states, and is interconnected by 46 arcs and considering the \textit{Intubation} variable as the target with three distinct class values. It is often regarded as a prototypical example of a BN employed to represent real-world scenarios.
\paragraph{Real world dataset :}
We employed the COMPAS dataset from ~\cite{angwin2016machine} (Correctional Offender Management Profiling for Alternative Sanctions), a widely used commercial algorithm employed by judges and parole officers to predict the likelihood of criminal defendants reoffending recidivism. Specifically, we worked with the \textit{compass\_violent\_parseda\_filt} dataset, considering the \textit{is\_recid} variable as the target with three distinct class values. After preprocessing, which involved removing ID, date, and case number-related features, the dataset was refined to include 27 features and 1 class variable with 3 labels.

The original dataset is divided into training and testing sets using an 80/20 split ratio, allocating 80\% of the data for training and 20\% for testing purposes. This split is performed exclusively on the dataset with preprocessed features. To maintain consistency and fairness, the datasets underwent preprocessing steps as outlined for each dataset. Notably, an additional preprocessing step was applied exclusively to the LaPLACE method, involving the discretization of continuous value features into 20 bins to align the data with the BN framework inherent to LaPLACE. 

Perturbed data integration contains the deliberate introduction of controlled variations or noise into testing instances. This procedure begins with the computation of attribute statistics from training data, followed by the generation of random values adhering to the distribution of each variable. These generated random values are subsequently employed to create perturbed data while maintaining data range limitations, thereby ensuring data consistency and enabling the assessment of model resilience. Subsequently, the original model (random forest) is leveraged to predict the target variable $T$ within the 5000 perturbed data in this experiment.

In the case of LIME and SHAP explanations, our approach involved selecting the top 5 variables as their respective explanations. These selected variables were then utilized for the purpose of comparison with our LaPLACE explainer, which averaged over 100 runs. 

\begin{table}[h]
\begin{center}
    \caption{Summary of datasets after preprocessing}\label{tab:data}
        \begin{tabular}{|l|c|c|c|c|}  
            \hline
             \multirow{3}{*}{Datasets} & \multicolumn{4}{c|}{Number of} \\
			 \cline{2-5}
            	& features  & \makecell{ Distinct \\ Values } & instances & labels  \\
            \hline
             ALARM & 36  & 2-4  & 10,000 & 3 \\    
             COMPAS & 27 & 2-20  &  18,316 &  3 \\
            \hline
        \end{tabular}
    \end{center}
\end{table}
\subsection{Evaluation metrics}
\paragraph{Local accuracy :} We compared the efficacy of explanations from our explainer with well-established methods, LIME and SHAP, across widely utilized classifiers such as Bayesian network (\textbf{BN}), random forests (\textbf{RF}), and support vector machines (\textbf{SVM}) on the dataset by using only explained features from their explanations. The objective was to quantitatively assess the faithfulness and correctness of information conveyed by the explained feature subset. Within this framework, we present the average weighted F1 score for reliable predictions on 20\% testing dataset across various explanation methods, as summarized in Table~\ref{tab:exp1}. 

\paragraph{Consistency :} We put all of these explanations into a unified feature set and proceeded to evaluate their concordance by quantifying the entropy of the combined set. Ideally, a lower value signifies a better agreement. This metric of consistency, known as entropy-based measurement, allows us to measure the alignment among the various explanations, in Table~\ref{tab:exp2}.

\begin{table}[t]
	\begin{center}
		
		\begin{tabular}{|cc|ccc|c|}
			\hline
			
			\multirow{3}{*}{Exp.} &\multirow{3}{*}{Datasets} & \multicolumn{3}{c|}{Classifiers}& \multicolumn{1}{c|}{} \\
			\cline{3-5}
			& &   RF & BN & SVM & \makecell{ Avg. \# \\ Explained  \\ Features}      \\
			\hline
			\multirow{3}{*}{LIME} & ALARM & \makecell{0.945 \\$\pm$ \\ 0.029} & \makecell{0.921 \\$\pm$ \\ 0.037}  & \makecell{0.933 \\$\pm$ \\ 0.034}  & \multirow{2}{*}{\makecell{5\\(predefined)}}  \\  
			\cline{3-5}
			&COMPAS & \makecell{0.953 \\$\pm$ \\ 0.094}  & \makecell{0.0.952 \\$\pm$ \\ 0.098} & \makecell{0.952 \\$\pm$ \\ 0.099} &\\

			\hline
			\multirow{3}{*}{SHAP} & ALARM & \makecell{0.959 \\$\pm$ \\ 0.017}  & \makecell{0.951 \\$\pm$ \\ 0.015} & \makecell{0.957 \\$\pm$ \\ 0.018} & \multirow{2}{*}{\makecell{5\\(predefined)}}   \\  
			\cline{3-5}
			&COMPAS & \makecell{0.932 \\$\pm$ \\ 0.016} & \makecell{0.932 \\$\pm$ \\ 0.016}  &  \makecell{0.932 \\$\pm$ \\ 0.016} &\\
			 
			\hline
            \multirow{3}{*}{\makecell{La  \\ PLACE}} & ALARM & \makecell{\textbf{0.981} \\$\pm$ \\ 0.002} &  \makecell{\textbf{0.965} \\$\pm$ \\ 0.002} &  \makecell{\textbf{0.973} \\$\pm$ \\ 0.000} & 5.79   \\  
			\cline{3-5}
			&COMPAS & \makecell{\textbf{1.000} \\$\pm$ \\ 0.000} & \makecell{\textbf{1.000} \\$\pm$ \\ 0.000} & \makecell{\textbf{1.000} \\$\pm$ \\ 0.000} & 4.70\\
			
		\hline

		\end{tabular}

	\end{center}	
	\caption{Local accuracy : average and standard deviation of weighted F1 on 100 explanations validated with (20\%) testing datasets on a collection of classifiers.}\label{tab:exp1}
\end{table}

\begin{table}[h]
\begin{center}
    
        \begin{tabular}{|l|cc|}  
            \hline
             \multirow{3}{*}{Explainer} & \multicolumn{2}{c|}{Datasets} \\
			\cline{2-3}
            	& ALARM  & COMPAS   \\
            \hline
             LIME & 3.78  & 3.79    \\    
             SHAP & 4.23 & 3.93   \\
             LaPLACE & \textbf{2.72} & \textbf{2.31}  \\

            \hline
        \end{tabular}
    \end{center}
    \caption{Consistency (entropy scores) of explanations across all tested datasets.}\label{tab:exp2}
\end{table}

\subsection{Discussion}

In our endeavor to provide explanations for individual predictions, we conducted a comprehensive comparative analysis involving LaPLACE, LIME, and SHAP. According to our study, the following observations can be made:
\begin{enumerate}
    \item When assessing classification performance through RF, BN, and SVM models and considering the explanations provided by LIME and SHAP, we observe notable variability across diverse scenarios. Specifically, in both datasets, LaPLACE yield higher average weighted F1 scores compared to LIME and SHAP. Importantly, it is evident that both LIME and SHAP show greater standard deviations in their performance when compared with LaPLACE across all datasets in Table~\ref{tab:exp1}. This observation highlights that explanations derived from LaPLACE contribute to enhanced prediction stability and faithful explanation in comparison to the other two methods.
    
    \item LaPLACE-Explainer exhibited a higher degree of concordance among their explanations, evidenced by their ability to accurately reflect  associated features to the predicted variable with better entropy scores compared to LIME and SHAP in Table~\ref{tab:exp2}. This alignment underscores the effectiveness of LaPLACE explanations presenting consistent explanations, enhancing their interpretability and utility in real-world applications. 
\end{enumerate}

 While LIME and SHAP provide explanations centered on significant feature values, a notable challenge arises in determining the optimal number of top variables ($N$) to serve as explanations. This task lacks a definitive solution and varies across datasets due to the uncertainty of necessary features. In contrast, LaPLACE introduces autonomously identifying the optimal subset of pivotal features essential for elucidating the target variable. This adaptability eliminates the need to specify a predetermined $N$ value, enhancing the explanation process and robustness of our model.

An additional noteworthy strength of the LaPLACE explainer lies in its ability to pinpoint fairness-related concerns within predictions. The LaPLACE explanations excel in transparently highlighting the presence of sensitive attributes, such as race or gender, within the explanatory framework. This transparency becomes particularly valuable in systems involving sensitive attributes, enabling an intuitive comprehension of their impact, whether through direct causal relationships or indirect associations. Even in cases where explicit sensitive attributes are removed from the training data, the potential for indirect correlations remains, emphasizing the significance of their detection to mitigate potential machine bias. For instance, consider the single inclusion of such attributes as $\mathbf{Sp}(T)$ in the explained $\mathbf{MB}(T)$. However, the incorporation of sensitive attributes necessitates careful consideration, as it introduces the potential for algorithmic bias when explanations interact with human subjects. This indicates that a rigorous assessment is required to ensure that the original algorithm avoids inadvertently perpetuating or amplifying biases, safeguarding fairness and impartiality in decision-making contexts.

While LaPLACE-Explainer offers a range of significant advantages as detailed, it's essential to acknowledge that it entails a trade-off in terms of computational complexity and limitations,  such as the need for  high-quality data for reliable conditional independence (CI) tests and the computational demands of learning the causal structure. The time complexity of CI test is $O(N (|\mathbf{Z}|+2))$, where each dataset point is examined just once to construct all non-zero entries in the contingency table. It relates to the number of variables in the test. The time needed is proportional to the sum of the cardinalities of ${X, Y}$ and the conditioning set $\mathbf{Z}$, given as $|{X, Y} + \mathbf{Z}| = 2 + |\mathbf{Z}|$. The process of learning the causal structure and identifying the Markov blanket involves intensive calculations, demanding substantial computational resources and time. This price of computational complexity is a consideration that needs to be weighed against the benefits of the transparent and meaningful explanations it provides. 

\section{Conclusion}

The LaPLACE explainer offers a distinctive advantage by providing users with transparent cause-and-effect explanations that eliminate the need for predefining a specific top ($N$) number of explanatory features. Its effectiveness is evidenced by the consistent integration of the Markov blanket of the target variable within explanations. However, this comes with the trade-off of high computational complexity. The LaPLACE-Explainer's ability to navigate attribute's causal relationships and construct comprehensive causal graphs demands intensive calculations, particularly in extensive datasets or high-dimensional feature spaces. This complexity underscores its commitment to providing profound insights, and while necessitating more computational resources, the resultant understanding and informed decision-making underscore its value. Another notable issue that we have not addressed in this work is the assessment of fairness. Fairness evaluation is a multifaceted challenge, as it involves accommodating diverse requirements across various domains and fairness-aware algorithms to find the trade-off between fairness requirements and performance.

Future research directions for our approach are promising. We aim to address challenges through potential parallelization or GPU utilization for improved efficiency. Additionally, we see the potential to broaden our approach to encompass diverse explanation models suitable for any type of machine learning model. This could involve exhaustive search techniques guided by various score functions or search constraints, inviting a comparative study involving domain experts, particularly in high-stakes fields like criminal justice and healthcare.

Moreover, the exploration of heuristic methods with polynomial time complexity is a promising avenue. This could enhance the scalability and applicability of the MB-based explanation induction process to complex real-world datasets with high-dimensional features. To overcome challenges, future research endeavors aspire to develop algorithms that strike a balance between scalability, time efficiency, data efficiency, causality, and trustworthiness. The ultimate aim is to achieve an equilibrium capable of effectively handling large-scale datasets, exploring the efficient search space, leveraging causal relationships, and delivering interpretable outcomes. Pursuing these research directions will advance our understanding of explanations for classifiers, contributing to more robust and interpretable machine learning systems.

\bibliography{aaai22}

\end{document}